\documentclass[11pt]{article}
\usepackage{colacl}

\title{Segregatory Coordination and Ellipsis in Text Generation}

\author{James Shaw \\ 
Dept.~of Computer Science\\
Columbia University\\
New York, NY 10027, USA\\
shaw@cs.columbia.edu 
}

\newcommand{\plandoc}{{\sc PLANDoc}}

\newcommand{\casper}{{\sc Casper}}

\begin{document}
\maketitle

\begin{abstract}
In this paper, we provide an account of how to generate
sentences with coordination constructions from clause-sized semantic
representations.  An algorithm is developed and various examples from
linguistic literature will be used to demonstrate that the algorithm
does its job well.
\end{abstract} 

\bibliographystyle{/u/peptic/shaw/papers/acl98/Final/acl}

\section{Introduction}

The linguistic literature has described numerous coordination 
phenomena \cite{gleitman65,ross67,neijt79,quirk85,vanoirsouw87,steedman90,pollard94,carpenter98}.
We will not address common problems associated with parsing, such as
disambiguation and construction of syntactic structures from a string.
Instead, we show how to generate sentences with complex
coordinate constructions starting from semantic representations.  We
have divided the process of generating coordination expressions into
two major tasks, identifying recurring elements in the conjoined
semantic structure and deleting redundant elements using syntactic
information.  Using this model, we are able to handle coordination
phenomenon uniformly, including difficult cases such as
non-constituent coordination.

In this paper, we are specifically interested in the generation of
segregatory coordination constructions.  In {\em segregatory}
coordination, the coordination of smaller units is logically
equivalent to coordination of clauses; for example, ``John likes Mary
and Nancy'' is logically equivalent to ``John likes Mary''
and ``John likes Nancy''.  Other similar conjunction
coordination phenomena, such as {\em combinatory} and {\em rhetorical}
coordination, are treated differently in text generation systems.
Since these constructions cannot be analyzed as separate clauses, we
will define them here, but will not describe them further in the
paper.  In combinatory coordination, the sentence ``Mary and Nancy are
sisters.'' is not equivalent to ``Mary is a sister.'' and ``Nancy is a
sister.''  The coordinator ``and'' sometimes
can function as a rhetorical marker as in ``The train sounded the
whistle and [then] departed the station.''\footnote{The string enclosed in
symbols [ and ] are deleted from the surface expression, but these
concepts exist in the semantic representation.}

To illustrate the common usage of coordination constructions, we will
use a system which generates reports 
describing how much work each employee has performed in an imaginary
supermarket human resource department.
Generating a separate sentence for each tuple in
the relational database would result in: ``John rearranged
cereals in Aisle 2 on Monday.  John rearranged candies in Aisle 2 on
Tuesday.''  A system capable of generating segregatory
coordination construction can produce a shorter sentence: ``John
rearranged cereals in Aisle 2 on Monday and candies on Tuesday.''

In the next section, we briefly describe the architecture of our
generation system and the modules that handle coordination
construction.  A comparison with related work in text generation is
presented in Section~\ref{sec:related}. 
In Section~\ref{sec:semrepsec}, we describe the
semantic representation used for coordination.  An algorithm for
carrying out segregatory coordination is provided in
Section~\ref{sec:algosec} with an example.  In
Section~\ref{sec:linguistsec}, we will analyze various examples taken
from linguistic literature and describe how they are handled by the
current algorithm.

\section{Generation Architecture}
\label{sec:archsec}

Traditional text generation systems contain a strategic and a tactical
component.  The strategic component determines what to say and the
order in which to say it while the tactical component determines how
to say it.  Even though the strategic component must first decide
which clauses potentially might be combined, it does not have access
to lexical and syntactic knowledge to perform clause combining as the
tactical component does.  We have implemented a sentence planner,
\casper\ ({\bf C}lause {\bf A}ggregation in {\bf S}entence {\bf
P}lann{\bf ER}), as the first module in the tactical component to
handle clause combining.  The main tasks of the sentence planner are
clause aggregation, sentence boundary determination and paraphrasing
decisions based on context \cite{wanner96,shaw95}.  
The output of the sentence planner is an ordered list of semantic
structures each of which can be realized as a sentence.  A lexical
chooser, based on a lexicon and the preferences specified from the
sentence planner, determines the lexical items to represent the
semantic concepts in the representation.  The lexicalized result is
then transformed into a syntactic structure and linearized into a
string using FUF/SURGE \cite{elhadad-phd,robin-phd}, a realization
component based on Functional Unification
Grammar \cite{halliday94,kay84}.

Though every component in the architecture contributes to the
generation of coordinate constructions, most of the coordination
actions take place in the sentence planner and the lexical chooser.
These two modules reflect the two main tasks of generating
coordination conjunction: the sentence planner identifies recurring
elements among the coordinated propositions, and the lexical chooser
determines which recurring elements to delete.  The reason for such a
division is that ellipsis depends on the sequential order of the
recurring elements at surface level.  This information is only
available after syntactic and lexical decisions have been made.  For
example, in ``On Monday, John rearranged cereals in Aisle 2 and
cookies in Aisle 4.'', the second time PP is deleted, but in
``John rearranged cereals in Aisle 2 and cookies in Aisle 4 on
Monday.'', the first time PP is deleted.\footnote{
The expanded first example is ``On Monday, John rearranged cereals in
Aisle 2 and [on Monday], [John] [rearranged] cookies in Aisle 4.''  
The expanded second example is ``John rearranged cereals in Aisle 2 [on
Monday] and [John] [rearranged] cookies in Aisle 4 on Monday.''}
\casper\ 
only marks the elements as recurring and let the lexical chooser make
deletion decisions later.  A more detailed description is provided in
Section~\ref{sec:algosec}.

\section{Related Work}
\label{sec:related}
Because sentences with coordination can express a lot of
information with fewer words, many text generation systems have
implemented the generation of coordination with various levels of
complexities.  In earlier systems such as EPICURE \cite{dale92c1},
sentences with conjunction are formed in the strategic component as
discourse-level optimizations.  Current systems handle aggregations
decisions including coordination and lexical aggregation, such as
transforming propositions into modifiers (adjectives, prepositional
phrases, or relative clauses), in a sentence planner
\cite{scott90,dalianis93,huang96,callaway97,shaw98a}.  Though other
systems have implemented coordination, their aggregation rules only
handle simple conjunction inside a syntactic structure, such as
subject, object, or predicate.  In contrast to these localized rules,
the staged algorithm used in \casper\ is global in the sense that it
tries to find the most concise coordination structures across all the
propositions.  In addition, a simple heuristic was proposed to
avoid generating overly complex and potentially ambiguous sentences as
a result of coordination.  \casper\ also systematically handles
ellipsis and coordination in prepositional clauses which were not
addressed before.  When multiple propositions are combined, the
sequential order of the propositions is an interesting issue.
\cite{dalianis93} proposed a domain specific ordering, such as
preferring a proposition with an animate subject to appear before a
proposition with an inanimate subject.  \casper\ sequentializes the
propositions according to an order that allows the most concise
coordination of propositions.

\section{The Semantic Representation}
\label{sec:semrepsec}

\casper\ uses a representation influenced by 
Lexical-Functional Grammar \cite{bresnan82} and Semantic
Structures \cite{jackendoff90}.  While it would have been natural to
use thematic roles proposed in Functional Grammar, because our
realization component, FUF/SURGE, uses them, these roles would add
more complexity into the coordination process.  One major task of
generating coordination expression is identifying identical elements
in the propositions being combined.  In Functional Grammar, different
processes have different names for their thematic roles (e.g., MENTAL
process has role SENSER for agent while INTENSIVE process has role
IDENTIFIED). As a result, identifying identical elements under various
thematic roles requires looking at the process first in order to
figure out which thematic roles should be checked for redundancy.
Compared to Lexical-Functional Grammar which uses the same feature
names, the thematic roles for Functional Grammar makes the
identifying task more complicated.

\begin{figure}
\small
\hrulefill
\vspace{-0.1in}
\begin{verbatim}
((pred ((pred c-lose) (type EVENT)
        (tense past)))
 (arg1 ((pred c-name) (type THING)
        (first-name ``John'')))
 (arg2 ((pred c-laptop) (type THING)
        (specific no)
        (mod ((pred c-expensive)
              (type ATTRIBUTE)))))
 (mod ((pred c-yesterday) 
       (type TIME))))
\end{verbatim}
\vspace{-0.15in}
\caption{Semantic representation for 
``John lost an expensive laptop yesterday.''}
\label{fig:semrep}
\normalsize
\vspace{-0.1in}
\hrulefill
\vspace{-0.1in}
\end{figure}

\begin{figure}
\small
\hrulefill
\vspace{-0.1in}
\begin{verbatim}
Al re-stocked milk in Aisle 5 on Monday.
Al re-stocked coffee in Aisle 2 on Monday.
Al re-stocked tea in Aisle 2 on Monday.
Al re-stocked bread in Aisle 3 on Friday.
\end{verbatim}
\vspace{-0.15in}
\caption{A sample of input semantic representations in surface form.}
\label{fig:stage0}
\normalsize
\vspace{-0.1in}
\hrulefill
\vspace{-0.1in}
\end{figure}

In our representation, the roles for each event or state are PRED,
ARG1, ARG2, ARG3, and MOD.  The slot PRED stores the verb concept.
Depending on the concept in PRED, ARG1, ARG2, and ARG3 can take on
different thematic roles, such as Actor, Beneficiary, and Goal in
``John gave Mary a red book yesterday.'' respectively.  The optional
slot MOD stores modifiers of the PRED.  It can have one or multiple
circumstantial elements, including MANNER, PLACE, or TIME.  Inside
each argument slot, it too has a MOD slot to store information such as
POSSESSOR or ATTRIBUTE.  An example of the semantic representation is
provided in Figure~\ref{fig:semrep}.

\section{Coordination Algorithm}
\label{sec:algosec}

We have divided the algorithm into four stages, where the
first three stages take place in the sentence planner and the last
stage takes place in the lexical chooser:
\begin{list}{}{\parsep 0in \topsep 0in \parskip 0in \itemsep .05cm}
\item [{\bf Stage 1:}] group propositions and order them according to their
similarities while satisfying pragmatic and contextual constraints.
\item [{\bf Stage 2:}] determine recurring elements in the ordered
propositions that will be combined.
\item [{\bf Stage 3:}] create a sentence boundary when the combined
clause reaches pre-determined thresholds.
\item [{\bf Stage 4:}] determine which recurring elements are
redundant and should be deleted.
\end{list}
In the following sections, we provide detail on each stage.  To
illustrate, we use the imaginary employee report
generation system for a human resource department in a supermarket.

\subsection{Group and Order Propositions}

It is desirable to group together propositions with similar elements
because these elements are likely to be inferable and thus redundant at
surface level and deleted.  There are many ways to group and order
propositions based on similarities.  For the propositions in
Figure~\ref{fig:stage0}, the semantic representations have the
following slots: PRED, ARG1, ARG2, MOD-PLACE, and MOD-TIME.
To identify which slot has the most similarity among its elements, we
calculate the number of distinct elements in each slot across the
propositions, which we call NDE (number of distinct elements).  For
the purpose of generating concise text, the system prefers to group
propositions which result in as many slots with NDE = 1 as possible.  For the
propositions in Figure~\ref{fig:stage0}, both NDEs of PRED and ARG1 are
1 because all the actions are ``re-stock'' and all the agents are
``Al''; the NDE for ARG2 is 4 because it contains 4 distinct elements:
``milk'', ``coffee'', ``tea'', and ``bread''; similarly, the NDE of
MOD-PLACE is 3; the NDE of MOD-TIME is 2 (``on Monday'' and ``on
Friday''). 

\begin{figure}
\small
\hrulefill
\vspace{-0.05in}
\begin{verbatim}
Al re-stocked coffee in Aisle 2 on Monday.
Al re-stocked tea in Aisle 2 on Monday.
Al re-stocked milk in Aisle 5 on Monday.
Al re-stocked bread in Aisle 3 on Friday.
\end{verbatim}
\vspace{-0.15in}
\caption{\footnotesize Propositions in surface form after Stage 1.}
\label{fig:stage1}
\normalsize
\vspace{-0.1in}
\hrulefill
\vspace{-0.15in}
\end{figure}

The algorithm re-orders the propositions by sorting the elements in
each slots using comparison operators which can determine that Monday
is smaller than Tuesday, or Aisle 2 is smaller than Aisle 4.  Starting
from the slots with largest NDE to the lowest, the algorithm re-orders
the propositions based on the elements of each particular slot.  In
this case, propositions will ordered according to their ARG2 first,
followed by MOD-PLACE, MOD-TIME, ARG1, and PRED.  The sorting process
will put similar propositions adjacent to each other as shown in
Figure~\ref{fig:stage1}.

\subsection{Identify Recurring Elements}

\begin{figure}
\footnotesize
\hrulefill
\vspace{-0.1in}
\begin{verbatim}
((pred c-and) (type LIST)
 (elts
    ~(((pred ((pred "re-stocked") (type EVENT)
              (status RECURRING)))
       (arg1 ((pred "Al") (TYPE THING)
              (status RECURRING)))
       (arg2 ((pred "tea") (type THING)))
       (mod  ((pred "on") (type TIME)
              (arg1 ((pred "Monday")
                     (type TIME-THING))))))
      ((pred ((pred "re-stocked") (type EVENT)
              (status RECURRING)))
       (arg1 ((pred "Al") (TYPE THING)
              (status RECURRING)))
       (arg2 ((pred "milk") (type THING)))
       (mod  ((pred "on") (type TIME)
              (arg1 ((pred "Friday")
                     (type TIME-THING)))))))))
\end{verbatim}
\vspace{-0.1in}
\caption{\small The simplified semantic representation
for ``Al re-stocked tea on Monday and milk on Friday.''  Note: $\sim$()
$\equiv$ a list.}
\label{fig:betterfd}
\normalsize
\vspace{-0.1in}
\hrulefill
\vspace{-0.15in}
\end{figure}

The current algorithm makes its decisions in a sequential order and it
combines only two propositions at any one time.
The {\em result proposition} is a semantic representation which
represents the result of combining the propositions.  
One task of the sentence planner is to find a way to combine the next
proposition in the ordered propositions into the resulting
proposition.  
In Stage 2, it is concerned with how many slots have distinct values and
which slots they are.  When multiple adjacent
propositions have only one slot with distinct elements, these
propositions are {\em 1-distinct}.  A special optimization can be
carried out between the 1-distinct propositions by conjoining their
distinct elements into a coordinate structure, such as conjoined
verbs, nouns, or adjectives.
McCawley \cite{mccawley81} described this phenomenon as {\em Conjunction
Reduction} -- ``whereby conjoined clauses that differ only in one item can
be replaced by a simple clause that involves conjoining that item.''
In our example, the first and second propositions are 1-distinct at
ARG2, and they are combined into a semantic structure representing
``Al re-stocked {\em coffee} and {\em tea} in Aisle 2 on Monday.''  If
the third proposition is 1-distinct at ARG2 in respect to the result
proposition also, the element ``milk'' in ARG2 of the third
proposition would be similarly combined.  In the example, it is
not.  As a result, we cannot combine the third proposition using only
conjunction within a syntactic structure.

When the next proposition and the result proposition have more than
one distinct slot or their 1-distinct slot is different from the
previous 1-distinct slot, the two propositions are said to be {\em
multiple-distinct}.
Our approach in combining multiple-distinct propositions is different
from previous linguistic analysis.  Instead of removing recurring
entities right away based on transformation or substitution, the
current system generates {\em every} conjoined multiple-distinct
proposition.  During the generation process of each conjoined clause,
the recurring elements might be prevented from appearing at the
surface level because the lexical chooser prevented the realization
component from generating any string for such redundant elements.  Our
multiple-distinct coordination produces what linguistics describes
as ellipsis and gapping.  
Figure~\ref{fig:betterfd} shows the result
combining two propositions that will result in ``Al re-stocked tea on
Monday and milk on Friday.'' 
Some readers might notice that PRED and ARG1 in both propositions are
marked as RECURRING but only subsequent recurring elements are deleted
at surface level.  The reason will be explained in
Section~\ref{subsec:delrecur}.

\subsection{Determine Sentence Boundary}

Unless combining the next proposition into the result proposition will
exceed the pre-determined parameters for the complexity of a sentence,
the algorithm will keep on combining more propositions into the result
proposition using 1-distinct or multiple-distinct coordination.  In
normal cases, the predefined parameter limits the number of
propositions conjoined by multiple-distinct coordination to two.  
In special cases
where the same slots across multiple
propositions are multiple-distinct, the pre-determined limit is
ignored.  By taking advantage of 
parallel structures, these propositions can be combined using
multiple-distinct procedures without making the coordinate structure
more difficult to understand.  For example, the sentence ``John took
aspirin on Monday, penicillin on Tuesday, and Tylenol on Wednesday.''
is long but quite understandable.  Similarly, conjoining a long list
of 3-distinct propositions produces understandable sentences too:
``John played tennis on Monday, drove to school on Tuesday, and won
the lottery on Wednesday.''  These constraints allow \casper\ to
produce sentences that are complex and contain a lot of information,
but they are also reasonably easy to understand.

\subsection{Delete Redundant Elements}
\label{subsec:delrecur}

Stage 4 handles ellipsis, one of the most difficult phenomena to
handle in syntax.  In the previous stages, elements that occur
more than once among the propositions are marked as RECURRING, but the
actual deletion decisions have not been made because \casper\ lacks
the necessary information.
The importance of the surface sequential order can be demonstrated by the
following example.  In the sentence "On Monday, Al re-stocked coffee
and [on Monday,] [Al] removed rotten milk.", the elements in MOD-TIME
delete forward (i.e. the subsequent occurrence of the identical
constituent disappears).  When MOD-TIME elements are realized at the
end of the clause, the same elements in MOD-TIME delete backward
(i.e. the antecedent occurrence of the identical constituent
disappears): ``Al re-stocked coffee [on Monday,] and [Al] removed
rotten milk on Monday.''  Our deletion algorithm is an extension to
the Directionality Constraint in  \cite{tai69}, which is based on
syntactic structure.  Instead, our algorithm uses the
sequential order of the recurring element for making deletion decisions.
In general, if a slot is realized at the front or medial
of a clause, the recurring elements in that slot delete forward.  In
the first example, MOD-TIME is realized as the front adverbial while
ARG1, ``Al'', appears in the middle of the clause, so elements in both
slots delete forward.  On the other hand, if a slot is realized
at the end position of a clause, the recurring elements in such slot
delete backward, as the MOD-TIME in second example.  The extended
directionality constraint also applies to conjoined premodifiers and
postmodifiers as well, as demonstrated by
``in Aisle 3 and [in Aisle] 4'', and ``at 3 [PM] and [at] 9 PM''.

Using the algorithm just described, the result of the supermarket
example is concise and easily understandable:
``Al re-stocked coffee and tea in Aisle 2 and milk in Aisle 5 on
Monday.  Al re-stocked bread in Aisle 3 on Friday.''  Further
discourse processing will replace the second ``Al'' with a pronoun
``he'', and the adverbial ``also'' may be inserted too.

\casper\ has been used in an upgraded version of
\plandoc \cite{mckeown-anlp94}, a robust, 
deployed system which generates reports for justifying the cost to 
the management in telecommunications domain.  Some of the current
output is shown in Figure~\ref{fig:pdocoutput}.  
In the figure, ``CSA'' is a location; ``Q1'' stands for first quarter;
``multiplexor'' and ``working-pair transfer'' are telecommunications
equipment.  The first example is a typical simple proposition in the
domain, which consists of PRED, ARG1, ARG2, MOD-PLACE, and MOD-TIME.
The second example shows 1-distinct coordination at MOD-PLACE, where
the second CSA been deleted.  The third example demonstrates
coordination of two propositions with multiple-distinct in MOD-PLACE
and MOD-TIME.  The fourth example shows multiple things: the ARG1
became plural in the first proposition because multiple placements
occurred as indicated by simple conjunction in MOD-PLACE; the
gapping of the PRED ``was projected'' in the second clause was based
on multiple-distinct coordination.  The last example demonstrates the
deletion of MOD-PLACE in the second proposition because it is located
at the front of the clause at surface level, so MOD-PLACE deletes
forward.
\begin{figure}
\small
\hrulefill
\vspace{-0.05in}
\begin{enumerate}
\item 
The Base Plan called for one new fiber activation at CSA 1061
in 1995 Q2.
\item
New 150mb\_mux multiplexor placements were projected at CSA 1160 and
1335 in 1995 Q2.
\item 
New 150mb\_mux multiplexors were placed at CSA 1178 in 1995 Q4 and at CSA 
1835 in 1997 Q1.
\item 
New 150mb\_mux multiplexor placements were projected at CSA 1160, 1335
and 1338 and one new 200mb\_mux multiplexor placement at CSA 1913b in 1995 Q2.
\item
At CSA 2113, the Base Plan called for 32 working-pair transfers in 
1997 Q1 and four working-pair transfers in 1997 Q2 and Q3.
\end{enumerate}
\vspace{-0.05in}
\caption{Text generated by \casper.}
\label{fig:pdocoutput}
\vspace{-0.1in}
\hrulefill
\normalsize
\vspace{-0.15in}
\end{figure}

\section{Linguistic Phenomenon}
\label{sec:linguistsec}

In this section, we take examples from various linguistic
literature \cite{quirk85,vanoirsouw87} and show how the algorithm
developed in Section~\ref{sec:algosec} generates them.  We also show
how the algorithm can generate sentences with non-constituent
coordination, which pose difficulties for most syntactic theories.

Coordination involves elements of equal syntactic status.  Linguists
have categorized coordination into {\em simple} and {\em complex}.
Simple coordination conjoins single clauses or clause constituents
while complex coordination involves multiple constituents.  For
example, the coordinate construction in ``John {\em finished his work}
and [John] {\em went home}.''  could be viewed as a single proposition
containing two coordinate VPs.  Based on our algorithm, the phenomenon
would be classified as a multiple-distinct coordination between two
clauses with deleted ARG1, ``John'', in the second clause.  In our
algorithm, the 1-distinct procedure can generate many simple
coordinations, including coordinate verbs, nouns, adjectives, PPs,
etc.  With simple extensions to the algorithm, clauses with relative
clauses could be combined and coordinated too.

Complex coordinations involving ellipsis and gapping are much
more challenging.  In multiple-distinct coordination, each conjoined
clause is generated, but recurring elements among the propositions are
deleted depending on the extended directionality constraints mentioned in
Subsection~\ref{subsec:delrecur}.  It works because it takes advantage
of the parallel structure at the surface level.

Van Oirsouw \cite{vanoirsouw87}, based on the literature on coordinate
deletion, identified a number of rules which result in deletion under
identity: Gapping, which deletes medial material; Right-Node-Raising (RNR),
which deletes identical right most constituents in a syntactic
tree; VP-deletion (VPD), which deletes identical verbs and handles
post-auxiliary deletion \cite{sag76}.
Conjunction Reduction (CR), which deletes identical right-most or
leftmost material.  He pointed out that these four rules reduce the
length of a coordination by deleting identical material, and they serve
no other purpose.  
We will describe how our algorithm handles the examples van Oirsouw
used in Figure~\ref{fig:oirsouw}.

\begin{figure}
\hrulefill
\begin{list}{}{\parsep 0in \topsep 0in \parskip 0in \itemsep .05cm}
\item [{\bf Gapping:}]  John ate fish and Bill $\phi$ rice.
\item [{\bf RNR:}] John caught $\phi$, and Mary killed the rabid
dog. 
\item [{\bf VPD:}] John sleeps, and Peter does $\phi$, too.
\item [{\bf CR1:}] John gave $\phi$ $\phi$, and Peter sold a
record to Sue.
\item [{\bf CR2:}] John gave a book to Mary and $\phi$ $\phi$
a record to Sue.
\end{list}
\vspace{-0.05in}
\caption{Four coordination rules for identity deletion described by
van Oirsouw.}
\label{fig:oirsouw}
\vspace{-0.1in}
\hrulefill
\vspace{-0.15in}
\end{figure}

The algorithm described in Section~\ref{sec:algosec} can use the
multiple-distinct procedure to 
handle all the cases
except VPD.  
In the gapping example, the PRED
deletes forward.  In RNR, ARG2 deletes backward because it is positioned
at the end of the clause.  
In CR1, even though the medial slot ARG2
should delete forward, it deletes backward because it is considered at
the end position of a clause.  In this case, once ARG3 (the BENEFICIARY ``to
Sue'') deletes backward, ARG2 is at the end position of a 
clause.  This process does require more intelligent processing in the
lexical chooser, but it is not difficult.
In CR2, it is straight forward to delete forward because both
ARG1 and PRED are medial.  The current algorithm does not address
VPD.  For such a sentence, the system would have generated ``John and
Peter slept'' using 1-distinct.

Non-constituent coordination phenomena, the coordination of elements
that are not of equal syntactic status, are challenging for syntactic
theories.  The following non-constituent coordination can be
explained nicely with the multiple-distinct procedure.
In the sentence, ``The spy {\em was in his forties}, {\em of average
build}, and {\em spoke with a slightly foreign accent}.'', 
the coordinated constituents are VP, PP, and VP.
Based on our analysis,
the sentence could be generated by combining the first two clauses
using the
1-distinct procedure, and the third clause is combined using the
multiple-distinct procedure, with ARG1 (``the spy'') deleted forward.
\begin{quote}
The spy was in his forties, [the spy] [was] of average build, and [the spy]
spoke with a slightly foreign accent.
\end{quote}

\section{Conclusion}

By separating the generation of coordination constructions into two
tasks -- identifying recurring elements and deleting redundant elements
based on the extended directionality constraints, we are able to handle
many coordination constructions correctly, including
non-constituent coordinations.  Through numerous examples, we have
shown how our algorithm can generate complex coordinate
constructions from clause-sized semantic representations.  Both the
representation and the algorithm have been implemented and used in two
different text generation systems \cite{mckeown-anlp94,mckeown97}.

\section{Acknowledgments}

This work is supported by DARPA Contract DAAL01-94-K-0119, the
Columbia University Center for Advanced Technology in High Performance
Computing and Communications in Healthcare (funded by the New York
State Science and Technology Foundation) and NSF Grants GER-90-2406.

\end{document}